\documentclass[journal]{IEEEtran}

\ifCLASSINFOpdf
\else
\fi

\usepackage{graphicx}
\usepackage{amsmath}
\usepackage{amsthm}
\usepackage{booktabs}
\usepackage{algorithm}
\usepackage{algorithmic}
\usepackage{subfig,amssymb,multirow}
\usepackage{array}
\usepackage{url,soul,xcolor}

\usepackage[pagebackref=true,breaklinks=true,colorlinks,bookmarks=false]{hyperref}

\newcommand{\R}{\mathbb{R}}
\newcommand{\angstrom}{\text{\normalfont\AA}}

\newtheorem{theorem}{Theorem}

\begin{document}
\title{Symmetry Structured Convolutional Neural Networks}
%
%
%

\author{Kehelwala~Dewage~Gayan~Maduranga$^{1,*,\dag}$,
    Vasily~Zadorozhnyy$^{2,*}$,
    ~and~Qiang~Ye$^{2,\ddag}$\\
    $^{1}$ Department of Applied and Computational Mathematics and Statistics, University of Notre Dame\\
    $^{2}$ Department of Mathematics, University of Kentucky\\
    $^{1}$ {\tt\small kmaduran@nd.edu} \quad $^{2}$ {\tt\small \{vasily.zadorozhnyy, qye3\}@uky.edu}  
\thanks{$^{*}$ These authors contributed equally}
\thanks{$^{\dag}$ Work done while at the University of Kentucky}
\thanks{$^{\ddag}$ Research supported in part by NSF under grant DMS-1821144}
}
\maketitle

\begin{abstract}
    We consider Convolutional Neural Networks (CNNs) with 2D structured features that are symmetric in the spatial dimensions. Such networks arise in modeling pairwise relationships for a sequential recommendation problem, as well as secondary structure inference problems of RNA and protein sequences. We develop a CNN architecture that generates and preserves the symmetry structure in the network's convolutional layers. We present parameterizations for the convolutional kernels that produce update rules to maintain symmetry throughout the training. We apply this architecture to the sequential recommendation problem, the RNA secondary structure inference problem, and the protein contact map prediction problem, showing that the symmetric structured networks produce improved results using fewer numbers of machine parameters.
\end{abstract}


%
\IEEEpeerreviewmaketitle

\section{Introduction}

\IEEEPARstart{C}{onvolutional} Neural Networks (CNNs) were originally developed to efficiently model image data. They have been used in many other types of problems. In some applications of CNNs, unlike images, the hidden feature variables may possess certain structures such as being symmetric in the spatial dimensions. Here we are interested in architectural variations of CNNs that can produce symmetry structured features to improve performance for such problems with reduced computational time and memory costs. 

Consider for example features that describe some mutual interactions of subjects arranged in a 1D sequence. Such interaction is described by a matrix that is symmetric. Specifically, from a 1D sequential input, we want to produce a 2D matrix with the $(i,j)$ entry describing the interaction of subjects $i$ and $j$. In this case, the $(i,j)$ and $(j, i)$ entries describe the same mutual interaction of the object $i$ and $j$ and thus should be equal. One way of deriving a 2D interaction feature from a 1D sequence of $n$ subjects is to form a self-Cartesian of the 1D sequence, which can then be processed using a CNN. The self-Cartesian product is not symmetric in space and traditional CNNs do not produce symmetric features from the self-Cartesian input. Even if the input to a convolution layer is  symmetric, the convolution does not preserve the symmetry. For this type of application, only the lower triangular parts of feature maps are usually used to model the interactions while the remaining parts are discarded. For better interpretability and efficiency, it is therefore desirable to modify the traditional CNN architectures so that symmetric features can be generated and preserved.

In this paper, we address these problems by focusing on some applications that may  benefit from the network's symmetric structure. We propose {\em symmetry generating} convolutional layers 
that result in feature maps that are symmetric in the spatial dimensions. We show that the symmetry can improve the overall performance and training of the network. We also derive update rules for symmetry generating 
networks that can be used in a backpropagation algorithm. We employ our proposed symmetry structured CNNs in three application problems: the sequential recommendation problem~\cite{yan2019cosrec}, the RNA secondary structure prediction problem~\cite{willmott_2018}, and the Protein Contact map prediction problem~\cite{Wang_17AccurateDeNovo}. Our experimental results illustrate that  imposing the symmetric structure leads to an increase in the network's prediction capability, while reducing network parameters and hence the computational and memory costs. Our code for Sequential Recommendation problem is available at \href{https://github.com/vasily789/Symmetry-Structured-Convolutional-Neural-Networks}{https://github.com/vasily789/Symmetry-Structured-Convolutional-Neural-Networks}, and the code for RNA secondary structure inference problem is avaialable at \href{https://github.com/Gayan225/Symmetry-Structured-Convolutional-Neural-Networks}{https://github.com/Gayan225/Symmetry-Structured-Convolutional-Neural-Networks}.

\section{Related Work}
\begin{figure*}[t]
    \centering
    \includegraphics[width=\textwidth]{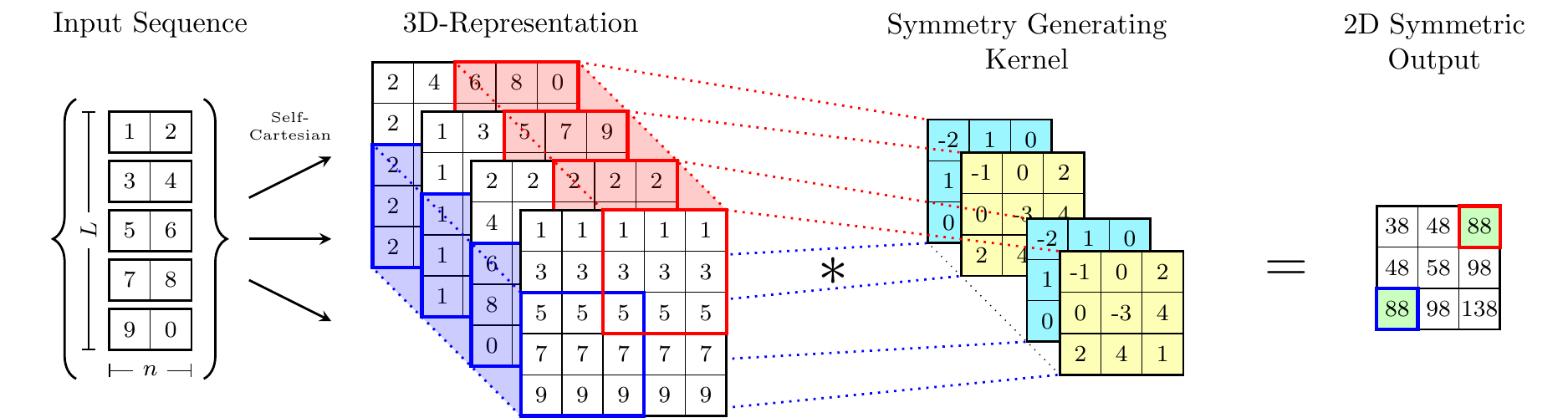}
    \caption{Illustration of a symmetry generating convolutional layer for one output channel: from an input of a vector sequence of dimension $L$, a self-Cartesian product produces a 3D $L\times L \times 2n$ representation. For one output channel, a $C\times C \times 2n$ ($C=3$ here) kernel has a symmetry in the channel direction and, for each channel, is symmetric in the spatial dimensions. This results in output features that are symmetric.
    }
    \label{fig:symmetry_generating_kernel}
\end{figure*}
CNNs with rotation and shift invariant  kernels were proposed in \cite{LO19951201} for medical image pattern recognition. The same group developed two other CNN systems with different kernel structures: the CNN with wavelet kernels(CNN/WK)~\cite{Lo95_wavelet}, and the CNN with circular kernels~\cite{LoSH98}. In the CNN/WK, the network forced each updated convolution kernel to be orthonormal, and thus, features selected on the transform domain are linearly independent. Hence, the fully connected layers in the classification level of CNN can perform more effectively. Other modifications of the CNN structure, the kernel structure, and the symmetric structure in CNNs in different aspects~\cite{Losh2002,DeepSymNet14NIPS2014_5424} have been proposed. More recently, rotation invariant CNNs based on Fourier Transform for a CNN was proposed by~\cite{Marcos_2016,chidester2018rotation}.

The transformationally identical CNN (TI-CNN) \cite{Lo_18} intends to exploit families of transformationally identical vectors that can make the CNN produce a quantitatively identical result through a series of processes in the CNN when a transformation of the input does not involve interpolation. TI-CNN uses a dihedral symmetric group structure for the kernel composed of reflection and $90^{\circ}$ rotation. The group-equivariant CNNs (G-CNNs)~\cite{CohenW16} generalizes the convolution operation of CNN from summing  over the spacial translations to an operation called G-correlation that sums over all transformations of  a symmetry group G. Such convolutions can capture features invariant under the symmetry group. Although the transpose operation may be included in such a symmetry group, these works are different from ours in that they do not attempt to produce symmetric output features, rather they compute features that detect possible symmetry invariance in the input.

Structure-aware convolution~\cite{SACNNNIPS2018_7287} (SACNNs) has been developed to eliminate a  limitation of CNNs in handling non-Euclidean structured data, such as the traffic flow data on traffic networks, the relational data on social networks, and the active data on molecule structure networks. SACNNs use structure-aware convolution in which a single shareable filter suffices to aggregate local inputs with diverse topological structures. For this purpose, SACNN generalize the classical filters to univariate functions that can be effectively and efficiently parameterized under the guidance of the function approximation theory and introduce local structure representations to quantificationally encode topological structures.

In \cite{DBLP_journals_corr_abs_1805_09421}, four different levels of symmetry in convolutional kernels are introduced for image applications as a regularizer to improve generalizations. They constrain the convolution kernels to be invariant under horizontal flips,  horizontal and vertical flips, as well as those plus 90$^{\circ}$ rotation. The feature maps of their networks do not necessarily possess a similar symmetry.  Our work differs from theirs in that our symmetry-structured networks generate and preserve features that are symmetric in the spacial dimensions and the symmetry we require of the kernel  is in the spacial dimensions  (i.e. the usual matrix symmetry) and  the channel dimension.   

\section{Symmetry Structured CNN}\label{SSCNN}

We are interested in the structure with the feature of the convolution layer  symmetric in the spacial dimension. Namely a convolution layer feature $\textbf{Z} \in \mathbb{R}^{n\times n \times c}$ is said to be symmetric if  $\textbf{Z}_{i,j,:}=\textbf{Z}_{j,i,:}$ for all  $i, j$. 
This kind of structure appears in the CosRec architecture~\cite{yan2019cosrec} pairwise embedding structure in the Sequential Recommendation Problem, RNA secondary structure inference problem~\cite{willmott_2018}, and protein contact map prediction problem~\cite{Wang_17AccurateDeNovo}. 
We introduce a symmetry generating kernel and a symmetry preserving kernel that can be used in the CNNs and propose update rules to be used in the backpropagation algorithm. 


\subsection{Symmetry Generating Kernel}\label{sec:21}

We consider the problem of using CNNs to generate a 2D feature matrix  
describing some mutual interactions of subjects in a 1D sequence. 
In order to use CNNs on a  sequence of  inputs, we need to generate two-dimensional structures first. The most natural way of doing this is to use the self-Cartesian product of the 1D input feature. The self-Cartesian product of a 1D sequence  $x = (x^{(1)}, \ldots, x^{(L)}) \in \R^{n \times L}$ such that $x^{(\ell)}=\left[x_{k}^{{\ell}}\right]_{k=1}^{n} \in \R^n$, with itself is the  tensor  $y = \left[y_{i,j,k}\right]_{i,j,k = 1}^{L,L,2n} \in \R^{L \times L \times 2n}$ as defined by:
\begin{equation}
\label{eq_selfCatesian}
y_{i,j,k} = \left\{
\begin{array}{ll}
      x_{k}^{(i)} & \text{if} \quad 1 \leq k \leq n, \\
       x_{k-n}^{(j)} & \text{if} \quad n+1 \leq k \leq 2n.
\end{array}
\right.
\end{equation}
Namely, $y(i,j,:)$ is the $i$th and $j$th terms $x^{(i)}$ and $x^{(j)}$ stacked together.
Clearly, $y$ is not symmetric. 
We first present a convolutional kernel that can produce a symmetric output from the  self-Cartesian product $y$. We call this type of kernel a symmetry generating kernel. 

Consider a  convolutional layer that takes  the self-Cartesian product as the input. 
Let $C, 2n$, and $F$ be the kernel size, the number of input channels, and the number of output channels respectively. We say  $\textbf{W} \in \mathbb{R}^{C \times C \times 2n \times F}$ is a  symmetry generating kernel if  $\textbf{W}_{i,j,:,:} = \textbf{W}_{j,i,:,:}$ and  $\textbf{W}_{i,j, k,:} = \textbf{W}_{i,j,n+k,:}$ for any $i, j$ and any $k\le n$.

\begin{theorem}
\label{SymmetryGeneratingThm}
Let $y$ be the self-Cartesian product of the 1D sequential input $x$ of the shape $[L,n]$. 
Consider a convolutional layer with $\textbf{W} \in \mathbb{R}^{C \times C \times 2n \times F}$ as the kernel and $y$ as the input. If  $\textbf{W}$ is a symmetry generating kernel, then the convolution layer's output is symmetric. 
\end{theorem}

See Figure \ref{fig:symmetry_generating_kernel} for an illustration of such a convolution layer and see Appendix \ref{Appendix:A} for a proof.

To train a symmetry generating convolutional layer, we first initialize the kernel as a symmetry generating kernel. However, if we train the network as usual, say, by a gradient descent step: $\textbf{W}^{(k+1)} =\textbf{W}^{(k)}-\alpha \frac{\partial L}{\partial \textbf{W}^{(k)}}$, the symmetry of $\textbf{W}^{(k)}$ is lost because $ \frac{\partial L}{\partial \textbf{W}^{(k)}}$ has no underlying structure of $\textbf{W}^{(k)}$. 
In order to keep the symmetric structure over the training, we force the symmetric property of a symmetry generating kernel $\textbf{W} \in \mathbb{R}^{C \times C \times 2n \times F }$ by parameterizing it with $\textbf{S} = [\textbf{S}_{i,j,k,f}] \in \mathbb{R}^{C \times C \times n \times F}$ where $i\geq j$ as: 

\begin{align}
    \label{eq_symgenKernel_init}
    \textbf{W}_{i,j,k,f}=\begin{cases} 
    \displaystyle \textbf{S}_{i,j,k,f} & \text{if } i\geq j,\, k \leq n,\\
    \displaystyle \textbf{S}_{j,i,k,f} & \text{if } i < j,\, k \leq n,\\
    \displaystyle \textbf{S}_{i,j,k-n,f} & \text{if } i \geq j,\, k > n,\\
    \displaystyle \textbf{S}_{j,i,k-n,f} & \text{if } i < j,\, k > n.
    \end{cases}
\end{align}

Namely, $\textbf{W} \in \mathbb{R}^{C \times C \times 2n \times F }$ is defined by  $\textbf{S}_{i,j,k,f}$ with $i\geq j$ and $1\le k \le n$. Then,  $\textbf{S}_{i,j,k,f}$ are trainable parameters and, during the training of CNNs, we use gradient descent on  $\textbf{S}_{i,j,k,f}$ and then update $\textbf{W} $. This also reduces the number of trainable parameters involved in training the kernel from $C^2 2n  F$ to $C (C+1) n F$.

Using $\textbf{S}_{i,j,k,f}$ as trainable parameters, we can compute the derivatives of a loss function from the derivatives with respect to $\textbf{W}$.
The following result gives the update rule that can be used in a backpropagation algorithm.

\begin{theorem}
\label{SymmetryGeneratingupdateThm}
Let $L$ be a differentiable loss function for a CNN with symmetry generating kernel $\textbf{W}$. i.e. $L=L(\textbf{W})$. Let $\textbf{W}$ be parameterized by $\textbf{S}_{i,j,k,f}$ with $i\geq j$ as in (\ref{eq_symgenKernel_init}). Then the gradient $\frac{\partial L}{\partial \textbf{S}} = \left[\frac{\partial L}{\partial \textbf{S}_{i,j,k,f}}\right]$ satisfies 

\begin{equation}
    \resizebox{.91\linewidth}{!}{$
    \label{eq_SymGenKernelUpdates1}
    \displaystyle\frac{\partial L}{\partial \textbf{S}_{i,j,k,f}}=\begin{cases} 
        \displaystyle \frac{\partial L}{\partial \textbf{W}_{i,j,k,f}} + \frac{\partial L}{\partial \textbf{W}_{j,i,k,f}} + \displaystyle \frac{\partial L}{\partial \textbf{W}_{i,j,k+n,f}}\\[8pt]
        \qquad +\displaystyle \frac{\partial L}{\partial \textbf{W}_{j,i,k+n,f}} \qquad \qquad  \text{ if } i > j,\\[8pt]
        \displaystyle \frac{\partial L}{\partial \textbf{W}_{i,i,k,f}} + \frac{\partial L}{\partial \textbf{W}_{i,i,k+n,f}} \quad \quad \text{if } i=j.
   \end{cases}
   $}
\end{equation}
\end{theorem}

See Appendix \ref{Appendix:A} for a proof. 

Thus, to train a symmetry generating kernel, we compute $\frac{\partial L}{\partial \textbf{W}}$, using the standard backpropagation algorithm and then, using $\frac{\partial L}{\partial \textbf{W}}$, we compute $\frac{\partial L}{\partial \textbf{S}}$ by Theorem \ref{SymmetryGeneratingupdateThm}. We then update $\textbf{S}$: $\textbf{S}^{(k+1)} =\textbf{S}^{(k)}-\alpha \frac{\partial L}{\partial \textbf{S}^{(k)}}$, where $\alpha$ is the learning rate, from which we obtain $\textbf{W}^{(k+1)}$.

In practice, a deep network needs more than one convolution layer. In subsequent convolution layers, the input is already symmetric. In order to maintain the symmetry in the output during the training, we use a symmetry preserving kernel $\textbf{Q} \in \mathbb{R}^{C \times C \times m \times F }$ where $\textbf{R} = [\textbf{R}_{i,j,k,f}] \in \mathbb{R}^{C \times C \times m \times F}$ with $i\geq j$: 

\begin{align}
    \label{eq_Kernel_init2}
    \textbf{Q}_{i,j,k,f}=\begin{cases} 
    \displaystyle \textbf{R}_{i,j,k,f} & \text{if } i\geq j,\\
    \displaystyle \textbf{R}_{j,i,k,f} & \text{if } i < j.
    \end{cases}
\end{align}
Namely, $\textbf{Q}$ is defined by $\textbf{R}_{i,j,k,f}$ with $i\geq j$. Then, in the training of CNNs, we update $\textbf{Q} $ through updates of $\textbf{R}_{i,j,k,f}$: 

\begin{align}
    \label{eq_SymPreserveKernel2Updates2}
    \frac{\partial L}{\partial \textbf{R}_{i,j,k,f}}=\begin{cases} 
         \displaystyle \frac{\partial L}{\partial \textbf{Q}_{i,j,k,f}}+ \frac{\partial L}{\partial \textbf{Q}_{j,i,k,f}} & \text{if } i > j, \\[8pt]
        \displaystyle \frac{\partial L}{\partial \textbf{Q}_{i,i,k,f}} & \text{if } i=j.
    \end{cases}
\end{align}

This also reduces  the number of trainable parameters from $C^2 m F$ to $C(C+1) m F/2$.

Using symmetry preserving kernels in the 2D convolution layers together with backpropagation updates from~(\ref{eq_SymPreserveKernel2Updates2}), we can maintain the symmetric structure in all feature maps. 

\begin{figure}
    \centering
    \subfloat[CNN]{{\includegraphics[width=3.5cm]{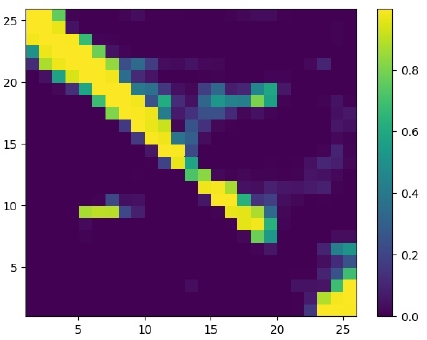} }}
    \qquad
    \subfloat[SCNN]{{\includegraphics[width=3.5cm]{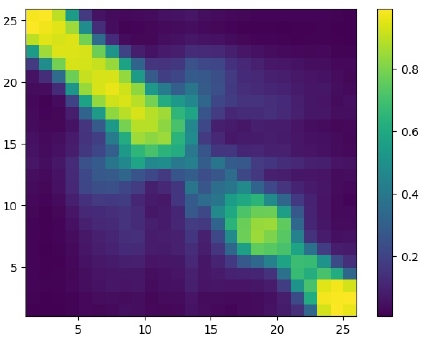} }}
    \caption{The heatmap plots of the activation maps for CNN and SCNN for the protein contact map experiment. }
    \label{fig:Heatmapplots}
\end{figure}

With the symmetry structured CNNs (SCNNs) the final output feature maps are symmetric. To illustrate the difference between SCNNs and CNNs, we show in
Figure~\ref{fig:Heatmapplots} the results of the output feature maps plotted in  heatmaps from the protein contact map prediction experiment in Section \ref{experiments}. Note that our SCNN, shown in Figure~\ref{fig:Heatmapplots}(b), produces a symmetric and smooth feature map that is physically more appealing, while the tradiational CNN, shown in Figure~\ref{fig:Heatmapplots}(a), leads to a nonsymmetric and less smooth feature map. As we will show in the experiments in Section \ref{experiments}, such a structure will benefit the performance of the networks.

\begin{table*}[t]
    \centering
    \begin{tabular}{c|c|c|c|c|c}
        \toprule
        \textbf{Dataset} & \textbf{Metric} & \textbf{GRU4Rec}$^*$ & \textbf{Caser}$^*$ & \textbf{CosRec}$^*$ & \textbf{SCosRec}(our) \\
        \midrule
        \multirow{7}{*}{\emph{ML-1M}}  & MAP           & 0.1440    & 0.1507    & 0.1883    & \textbf{0.1970} \\
                                        & Prec@1       & 0.2515    & 0.2502    & 0.3308    & \textbf{0.3458} \\
                                        & Prec@5       & 0.2146    & 0.2175    & 0.2831    & \textbf{0.2920} \\
                                        & Prec@10      & 0.1916    & 0.1991    & 0.2493    & \textbf{0.2586} \\
                                        & Recall@1     & 0.0153    & 0.0148    & 0.0202    & \textbf{0.0223} \\
                                        & Recall@5     & 0.0629    & 0.0632    & 0.0843    & \textbf{0.0895} \\
                                        & Recall@10    & 0.1093    & 0.1121    & 0.1438    & \textbf{0.1519} \\
        \midrule
        \multicolumn{4}{r|}{\# of trainable parameters:} & 1.983M & 1.730M\\
        \midrule
        \multirow{7}{*}{\emph{Gowalla}} & MAP          & 0.0580    & 0.0928    & 0.0980    & \textbf{0.1006} \\
                                        & Prec@1       & 0.1050    & 0.1961    & 0.2135    & \textbf{0.2171} \\
                                        & Prec@5       & 0.0721    & 0.1129    & 0.1190    & \textbf{0.1211} \\
                                        & Prec@10      & 0.0782    & 0.0571    & 0.0884    & \textbf{0.0898} \\
                                        & Recall@1     & 0.0155    & 0.0310    & 0.0337    & \textbf{0.0350} \\
                                        & Recall@5     & 0.0529    & 0.0845    & 0.0890    & \textbf{0.0920} \\
                                        & Recall@10    & 0.0826    & 0.1223    & 0.1305    & \textbf{0.1330} \\
        \midrule
        \multicolumn{4}{r|}{\# of trainable parameters:} & 5.641M & 5.383M\\
        \bottomrule
    \end{tabular}
    \caption{Sequential Recommendation problems: CosRec~\protect\cite{yan2019cosrec} vs. SCosRec. On both datasets, SCosRec outperforms the CosRec model in every metric. It also outperforms other neural network models such as GRU4Rec~\protect\cite{Hidasi2016SessionbasedRW} and Caser~\protect\cite{Cesar_18}; $^*$ - quoted from~\protect\cite{yan2019cosrec} and reproduced with our experiments.}
    \label{table_seq_recom}
\end{table*}


\subsection{Complexity and Initialization}
\label{sec:complexity} 
While our primary purpose of using the symmetry structured CNNs is to better model problems with underlying symmetric structure, our architectures also save about one half computational and memory cost in both training and inference. Consider the symmetry preserving kernel. As already mentioned, during training, a symmetry generating kernel has $C(C+1) n F$ trainable parameters, compared with $C^2 2n F$ of a full kernel. Furthermore, during inference, each feature map is symmetric in the spacial dimension requiring $L(L+1)m/2$ entries for input and  $L(L+1)F/2$ for output. This saves about half of the memory cost. Moreover, the convolution operation only needs to be computed for the upper triangular part of the output, which also approximately halves the computational cost. 

Finally, for the initialization, we suggest using half of the standard initialization such as Glorot~\cite{Glorot10} since each element of  $\textbf{S}$ and $\textbf{R}$ contributes twice in the $\textbf{W}$ and $\textbf{Q}$ respectively. So the half Glorot initialization is used in our experiments.

\section{Experiments}
\label{experiments}
In this section, we compare performance of a symmetry structured CNN  architecture with a corresponding traditional CNN. We mainly compare the two architectures by matching the architecture hyperparameters with respect to feature map dimensions and kernel dimensions. We study three applications of CNNs where the feature maps are naturally symmetric, which the traditional CNN architecture ignore.

\subsection{Sequential Recommendation}

Recommender systems have become a core technology in many applications. In sequential recommendation models, each user is represented as a sequence of items interacted with in the past and we aim to predict the next item or top $N$ items that a user will likely interact with in the near future. The order of interaction implies that sequential patterns play an essential role where more recent items in a sequence have a more significant impact on the next item. 
Some of early works on this problem include \cite{Hidasi2016SessionbasedRW,Cesar_18,yan2019cosrec}.

Our experiment was motivated by the  the most competitive model called CosRec  \cite{yan2019cosrec} that is based on CNNs. Having feature maps modeling interactions of items, it is well suited  for application of our symmetric kernels to the CNN layers. We compare the performance of our new model Symmetric CosRec (SCosRec) with the CosRec model.

The sequential recommendation problem can be formulated as follows.
Suppose we have a set of users $\mathcal{U} = {u_1,u_2,\ldots,u_{|\mathcal{U}|}}$ and a set of items $\mathcal{I} = {i_1,i_2,\ldots, i_{|\mathcal{I}|}}$. For each user $u \in \mathcal{U}$, given the sequence of previously interacted items $S^u = (S_1^u,\ldots S_{|S^u|}^u)$, $S_i^u \in \mathcal{I}$, we seek to predict the next item to match the user’s preferences. We follow the same setup given in the CosRec model that embeds the item matrix $E_{\mathcal{I}} \in \mathbb{R}^{|\mathcal{I}| \times d}$ and user matrix $E_{\mathcal{U}} \in \mathbb{R}^{|\mathcal{U}| \times d}$ , where $d$ is the latent dimensionality, $e_i$ and $e_u$ denote the $i$th and the $u$th rows in $E_{\mathcal{I}}$ and $E_{\mathcal{U}}$ respectively. Then for user $u$ at time step $t$, we retrieve the input embedding matrix $E_{u,t}^L \in \mathbb{R}^{L \times d}$ by looking up the previous L items $(S_{t-L}^u,\ldots S_{t-1}^u)$ in the item embedding matrix $E_{\mathcal{I}}$. Using pairwise encoding~\cite{yan2019cosrec}, CosRec  is a CNN model that creates a three-way tensor $T_{(u,t)}^L \in \mathbb{R}^{L \times L \times 2d}$ on top of the input embeddings $E_{u,t}^L$. The input structure of the three-way tensor $T_{(u,t)}^L$ can be incorporated with our symmetry generating and symmetry preserving CNN layers to obtain a symmetric  $T_{(u,t)}^L$ that better models the pairwise encoding. We call our model Symmetric CosRec or simply SCosRec.

We test our SCosRec model with hyperparameter matching architecture with the CosRec model on two standard benchmark datasets: \emph{MovieLens-1M (ML-1M)}~\cite{movielens} and \emph{Gowalla}~\cite{gowalla}. We use the \emph{MovieLens-1M (ML-1M)} version of a popular benchmark dataset for evaluating performance of collaborative filtering algorithms, and the \emph{Gowalla} dataset which is a location-based social networking website uses time and location information from check-ins made by users.

\emph{Evaluation metrics:} Results were evaluated in three top-$N$ metrics: Mean Average Precision (MAP), Precision@$N$, and Recall@$N$ with $N=1$, $5$, and $10$.

\emph{Implementation Details:} All experiments were implemented using Python 3.6.9 and PyTorch 1.1.0 on an NVIDIA Quadro P5000 GPU. Similarly to CosRec models, we used 2 \emph{symmetric} convolution blocks with 2 layers in each. The latent dimension $d$ is 50 and 100 for \emph{ML-1M} and \emph{Gowalla} datasets respectively. Markov order $L$ is 5, prediction of the next $T$ items is 3, the learning rate is $10^{-3}$, learning rate decay on plateau with reducing factor 0.15 and patience 3 with respect to the MAP metric, batch size is 512, negative sampling rate is 3, and dropout rate is 0.5.

Indeed, SCosRec models improvements are between $3-10\%$ on \emph{ML-1M} and $1-4\%$ on \emph{Gowalla} datasets compared to the original CosRec models while using less trainable parameters, see number of trainable parameters in Table \ref{table_seq_recom}. Results of our experiments are provided in Table \ref{table_seq_recom}.

\subsection{RNA Secondary Structure Inference}
\label{sec:rna}

\begin{figure}
    \centering
    \subfloat[Secondary structure]{{\includegraphics[width=3cm]{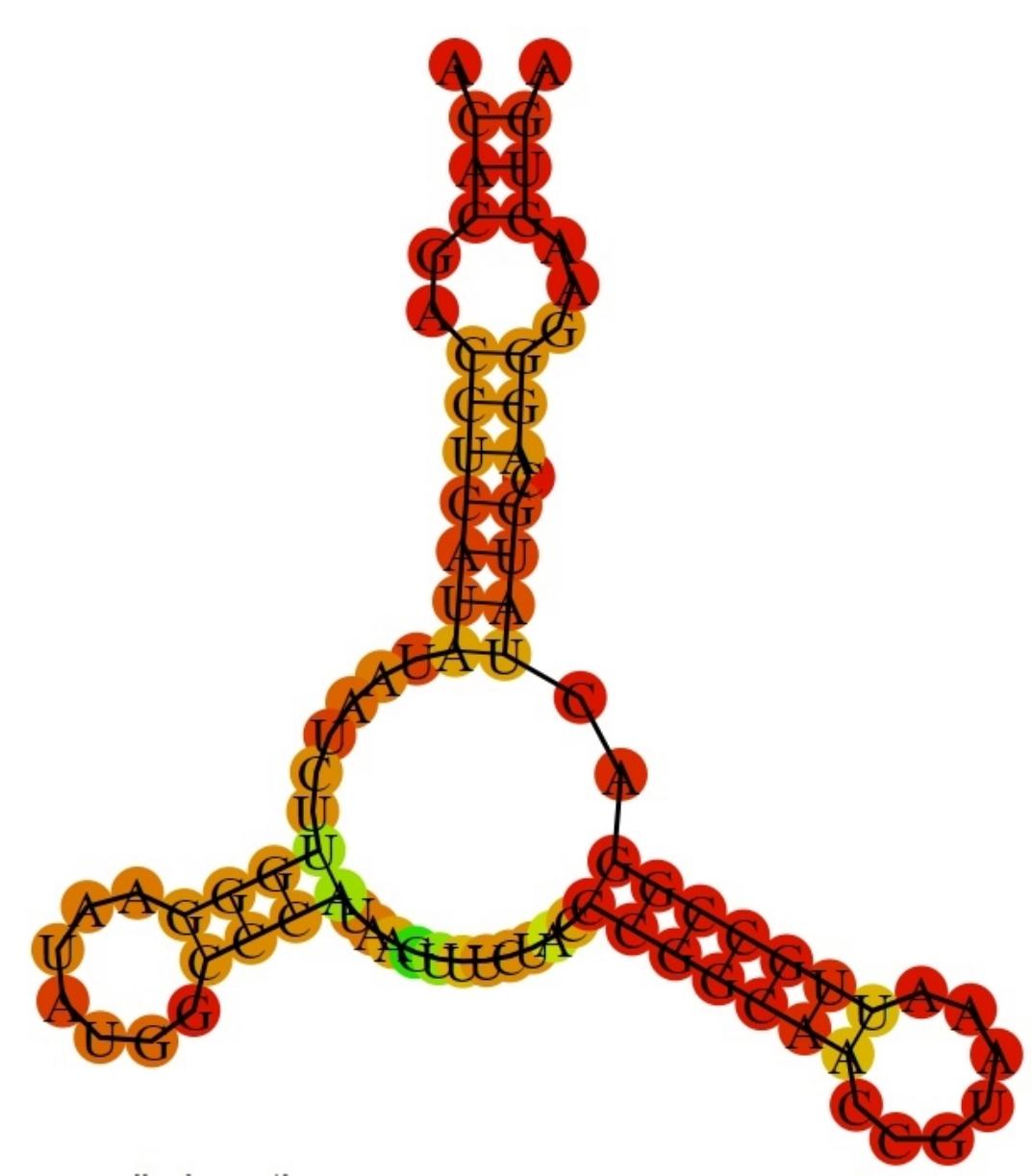} }}
    \qquad
    \subfloat[Matrix representation]{{\includegraphics[width=4cm]{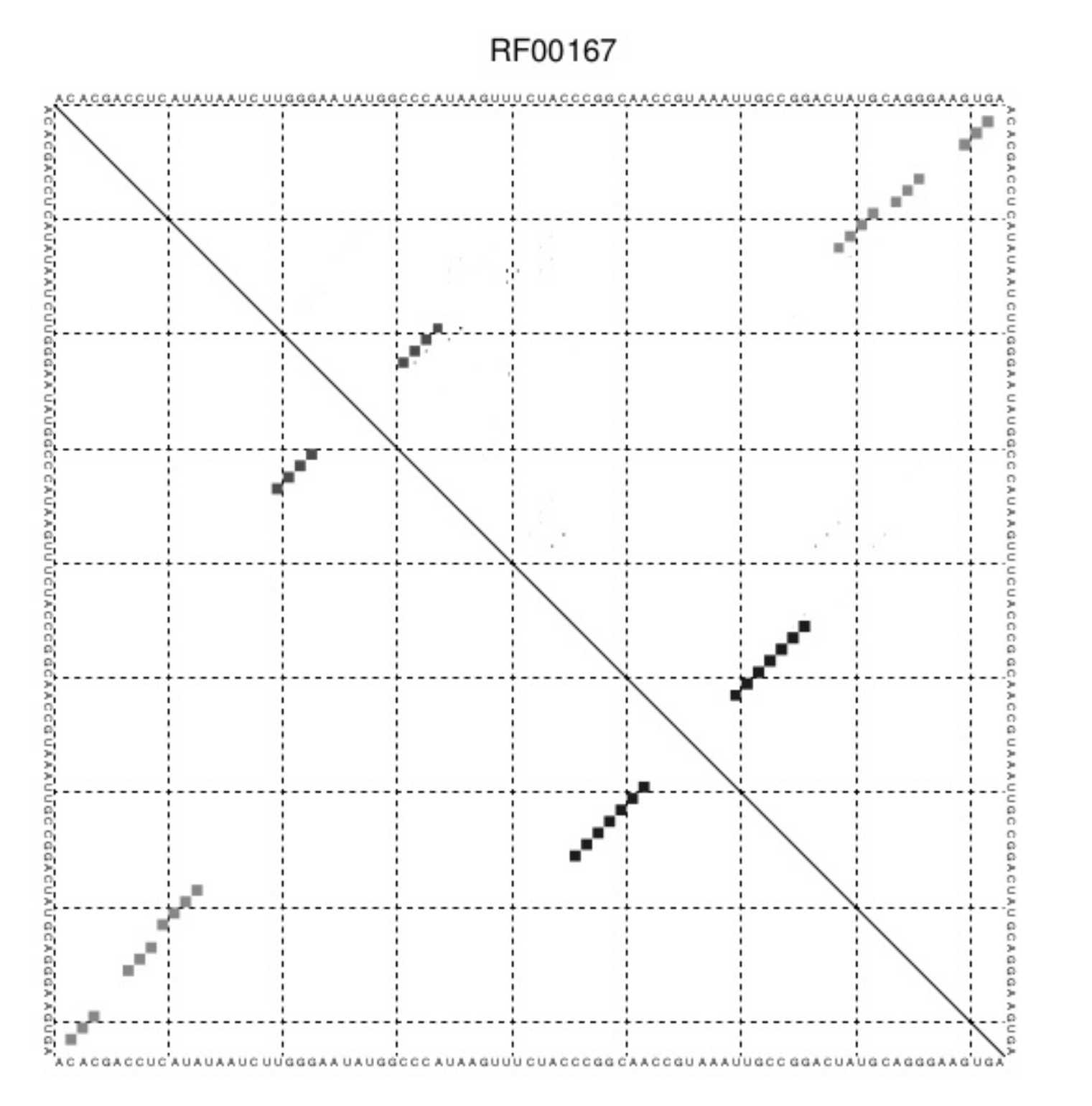} }}
    \caption{The secondary structures of  the purine riboswitch (Rfam RF00167)~\protect\cite{hofacker_stadler} with matrix representation. }
    \label{secondaryStructure2}
\end{figure}

Our second experiment is the problem of RNA secondary structure inference ~\cite{Gutell:2002kx}.
Given an RNA sequence $r= (r_1, \ldots, r_L)$ where each sequence element is a  nucleotide $r_i \in \{A,C,G,U,X\}$, we would like to find its native secondary structure $S$, which is primarily a list of base pairs of nucleotides in $r$. If there is a hydrogen bond between two nucleotides $r_i$ and $r_j$ in an RNA sequence, we say this bond forms a base pair $(i,j)$. 
Owing to these base pair bonds in RNA sequences, it creates a folded RNA structure that depends only on the sequence itself. 
Figure~\ref{secondaryStructure2} (a) gives an example of the secondary  structure for the RNA sequence of the purine riboswitch of length $83$. 
This secondary structure can be represented by a matrix $A \in \R^{L \times L} $ where $A^{(i,j)} = 1$ if $(i,j) \in S$ and $0$ otherwise, as in  Figure~\ref{secondaryStructure2} (b). 
Clearly, $A$ is a symmetric matrix.

\begin{figure}
    \centering
    \includegraphics[width=0.48\textwidth]{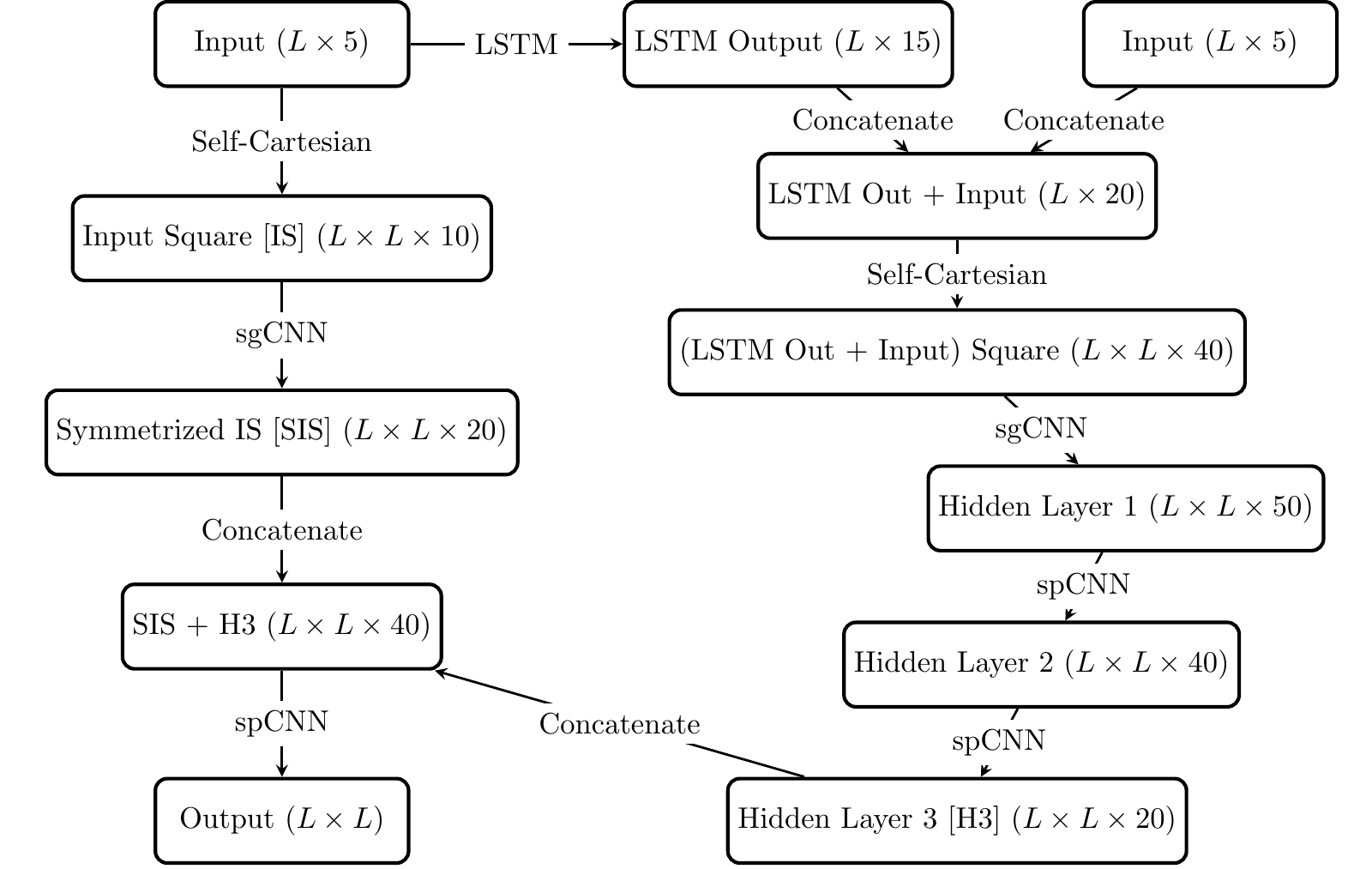}
    \caption{The symmetrized CNN  (SCNN) model for secondary structure inference that takes an input sequence ($L\times 5$) to produce an $L\times L$ symmetric matrix output. sgCNN and spCNN refer to symmetry generating  and symmetry preserving layers respectively.}
    \label{SymmetricStructureNetwork-diagram}
\end{figure}

The comparative sequence analysis~\cite{Gutell:2002kx} and the Nearest Neighbor Thermodynamics Model (NNTM)~\cite{Turner:2010uq} are the primary methods studied within the bioinformatics community for RNA structure inference.  A neural network model based on CNN  combined with LSTM was recently presented in \cite{willmott_2018} and was shown to significantly improve the NNTM results. This CNN model produces the secondary structure $A$ as output from an input of RNA sequence $r$. 
In order to obtain $A$, we  need to generate a 2D representation from the 1D sequence  $r= (r_1, \ldots, r_L)$ first, where each $r_i$ is encoded as a one-hot vector of dimension 5. This can be done using a Cartesian product of the sequence with itself; see~(\ref{eq_selfCatesian}). The output of this Cartesian product is further processed using a CNN to produce $A$. 
This CNN model, however, generates an unsymmetric feature matrix, from which only the upper triangular part is used to define the secondary structure and hence the loss function for training. The lower triangular part of the output is ignored and the network capacity is not fully utilized. The model also incorporates an LSTM network to obtain another sequential  feature that potentially captures global interaction of the sequence; see Figure~\ref{SymmetricStructureNetwork-diagram}. 

We implement this CNN model with either the symmetry generating kernel or symmetry preserving kernel in all convolutional layers to naturally produce a symmetric output that defines the secondary structure.  A detailed illustration of the model architecture is given in Figure~\ref{SymmetricStructureNetwork-diagram}. We test the original CNN and our SCNN models on the subset of 16S rRNA dataset from RNA STRAND~\cite{andronescu2008rna}, which contains 550 training, 28 validation, and 35 testing sequences.

\emph{Evaluation metrics:} To evaluate the performance, we compare a  predicted secondary structure $S_{\text{p}}$ with  the native structure $S_{\text{n}}$ in three metrics ~\cite{gardner2004comprehensive} as follows. A base pair $(i, j)$ is said to be {true positive} (TP) if it appears in both the predicted and the native structures, a false positive (FP) if it is in the predicted structure but not in the native structure, a false negative (FN) if it appears in the native structure and not in the predicted structure, and a true negative (TN) if it does not appear in either the predicted or the native structures.
Then, the {\em positive predictive value}  (PPV) is the proportion of true positives in the predicted structure, the {\em  sensitivity} is the fraction of true positives in the native structure, and the {\em accuracy} is 
the arithmetic mean of PPV and sensitivity. 

\emph{Implementation Details:} For the RNA direct secondary structure inference problem, all experiments were implemented using Python 3.7.6 and Tensorflow 2.0.0 on a V100 GPU. We modified the architecture given in the dissertation~\cite{willmott_2018}, which we will refer to as the CNN model for secondary structure inference to incorporate the symmetrical structure for the convolution layers, and we call it symmetrized CNN or SCNN for secondary structure inference. To train the parameters in the direct secondary structure prediction problem, we used the Adam optimizer~\cite{kingma2014adam} with an initial learning rate $\eta = 2 \times 10^{-3}$. We use batch size $10$ for the network training. We also use an $L2$ regularization term $10^{-5}$ and apply batch normalization~\cite{ioffe2015batch} after each convolutional layer. This has strong effects on the convergence speed and overall performance of the network training. We use the loss function as binary cross-entropy applied to each of the upper triangular entries $\frac{1}{2}L (L-1)$ of the network output $\hat Y \in \R^{L \times L}$. Note that we restrict that each nucleotide may pair with at most one other nucleotide. For the RNA sequence of length  $ L $, at most $ L $ of these entries will have positive labels. The machine sees many more negative training examples than positive examples. This imbalance will get worse as the sequence length increases. To remedy this situation, we use a weighted positive prediction term in the loss function using a small constant, and we found that either $3$ or $5$ improves our final accuracy.

As presented in Figure~\ref{SymmetricStructureNetwork-diagram}, we begin with an RNA sequence itself, a one-hot encoded to be of size $L \times 5$, with each class representing A, C, G, U, and X. We first pass the sequence through a bidirectional LSTM to generate an output. This output concatenated with the RNA sequence input gives a hidden layer of $L \times 20$. We then perform a self-Cartesian product, making arrays of size $L \times L \times 40$. These are passed through a symmetric generating convolution layer to obtain a symmetric hidden layer. We then run this hidden variable through several layers of symmetry preserving convolutional layers. We notice that using several different kernel sizes at each layer to capture features of varying sizes benifits the network performance. Each of the hidden convolutional layers applies a ReLU activation function, except the final layer's activation function is an elementwise sigmoid activation function. This makes the output $\hat Y \in \R^{L \times L}$ with $\hat Y^{(i,j)}$ representing the machine's prediction that the base pair $(i,j)$ is in the secondary structure $S$ of $r$.


Table \ref{16ssymmetric_Nonsymmetric1} presents the results on the 16S rRNA secondary structure inference task using the CNN model from \cite{willmott_2018} and our SCNN model. We consider hyperparameter matching models where we use the same kernel size and feature sizes as in CNN. 
In this case, we use fewer trainable parameters.
We note that our matching model improves the training, validation, and testing accuracy by $0.5\%, 1.9\%$, and $1.9\%$ with near $40\%$ fewer parameters. There is a similar increase in PPV and Sensitivity.

\begin{table}[!ht]
        \centering
        \begin{tabular}{l|l|c|c}
            \toprule
            \textbf{Set} & \textbf{Metric} & \textbf{CNN}$^*$ & \textbf{SCNN}\\
            \midrule
            \multirow{3}{*}{Training}
            & PPV & 0.933 & \textbf{0.934} \\%
            & Sen & 0.941 & \textbf{0.947}  \\%
            & Acc & 0.937 & \textbf{0.942}  \\ 
            \midrule
            \multirow{3}{*}{Validation}
            & PPV  & 0.899 & \textbf{0.916}  \\ 
            & Sen & 0.845 & \textbf{0.862}  \\ 
            & Acc & 0.872 & \textbf{0.891}  \\ 
            \midrule
            \multirow{3}{*}{Test}
            & PPV & 0.867 & \textbf{0.892} \\ 
            & Sen & 0.831 & \textbf{0.845}  \\
            & Acc & 0.849 & \textbf{0.868}  \\
            \midrule
            \multicolumn{2}{r|}{\# of trainable param.} & 387,798 & 239,198 \\
            \bottomrule
        \end{tabular}
        \caption{Results on the 16S rRNA secondary structure inference task using the CNN and our SCNN with hyperparameter matching model. $^*$ - quoted from~\protect\cite{willmott_2018} and reproduced.}
        \label{16ssymmetric_Nonsymmetric1}
\end{table}


\subsection{Protein Contact Map Prediction}
Our final experiment is the protein contact map prediction problem~\cite{Wang_17AccurateDeNovo}, which is similar to the RNA problem in \S \ref{sec:rna}.
Given a protein sequence $r= (r_1, \ldots, r_L)$ where each sequence element is one of $20$ different amino acids, we would like to find its contact map, represented by a binary two-dimensional matrix, that is defined from the distance between all possible amino acid residue pairs of a three-dimensional protein structure. For two residues $i$ and $j$, the $(i,j)$ element of the matrix is 1 if the two residues are closer than $8 \angstrom$, and 0 otherwise. 
For a protein sequence of length $L$, we use $L \times 25$ features which involve sequence profile and predicted structure, and $L \times L \times 3$ co-evolution and pairwise potential features to predict the contact map.

\begin{figure}[ht!]
\centering
\includegraphics[scale=0.5]{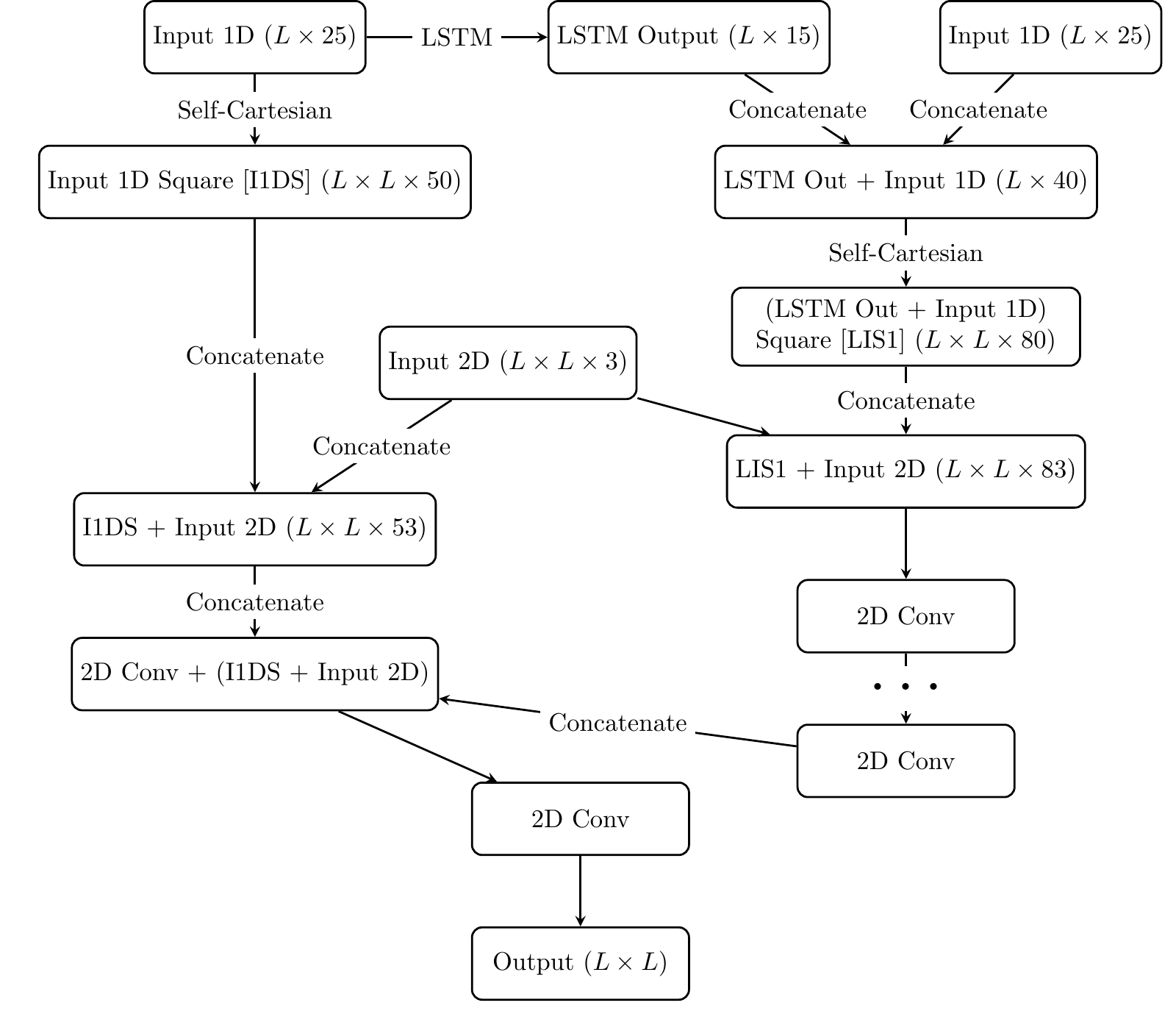}
\caption{The CNN model for protein contact map prediction that takes a 1D input sequence ($L\times 25$), and 2D input sequence ($L \times L \times 3$) to produce an $L\times L$ symmetric matrix output. 2D Conv refers to the two dimensional convolution layers.}
\label{Architecture1forcontactmap1}
\end{figure}

\begin{figure}[ht!]
\centering
\includegraphics[scale=0.5]{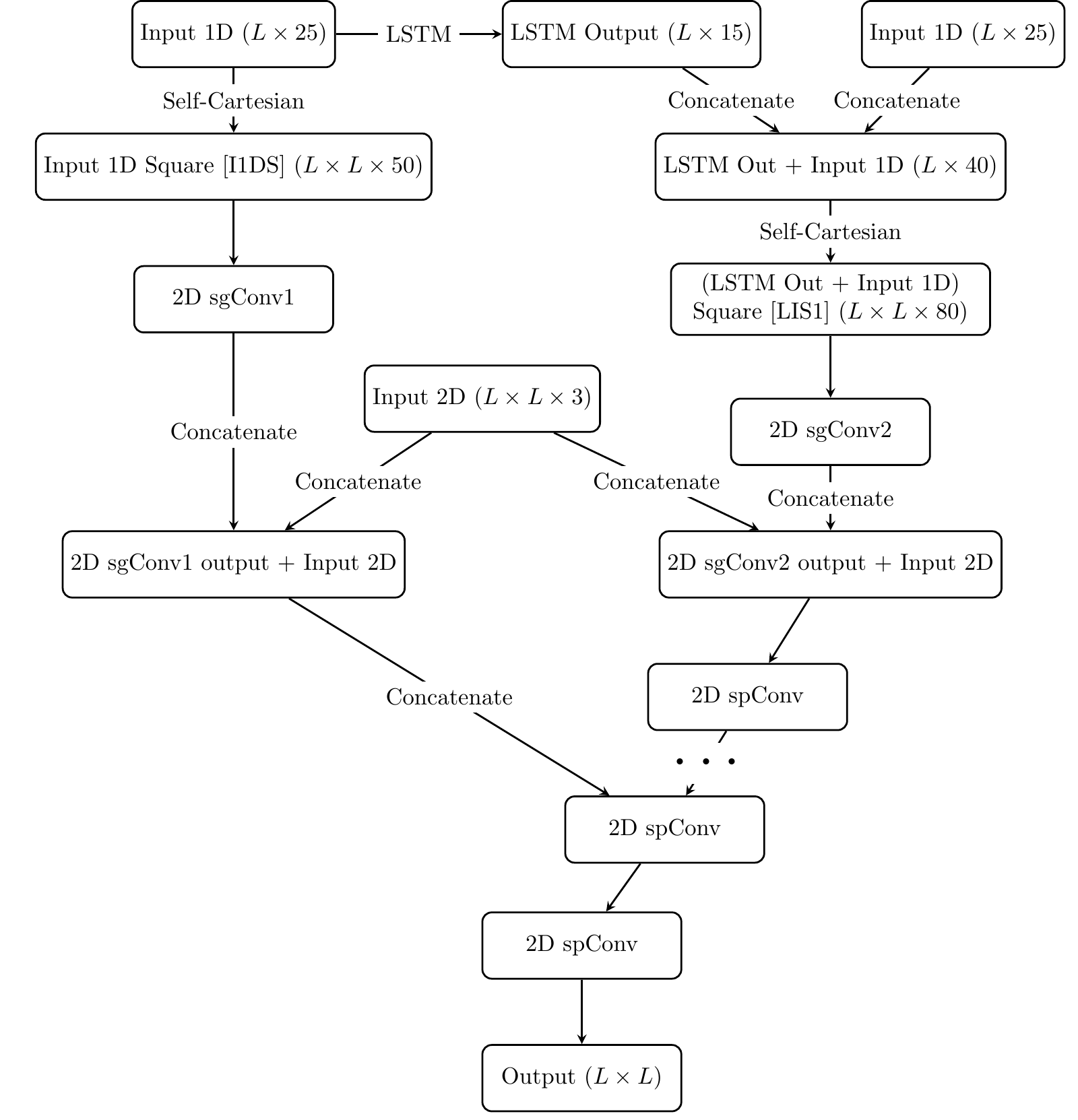}
\caption{The SCNN model for protein contact map prediction that takes 1D input sequence ($L\times 25$), and 2D input sequence ($L \times L \times 3$) to produce an $L\times L$ symmetric matrix output. Here sgConv stands for symmetry generating CNN while spConv for symmetry preserving CNN.}
\label{Architecture2forcontactmapSym}
\end{figure}

CNNs have been widely used in the protein contact map prediction problem  \cite{Wang_17AccurateDeNovo,Adhikari17}. 
As the final output is a symmetric matrix, we are interested in using our symmetry structured CNNs to improve the inference. 
As most existing works only provide an outline of their network architectures, 
we consider an architecture motivated by \cite{Wang_17AccurateDeNovo} and similar to the network (Figure~\ref{SymmetricStructureNetwork-diagram}) for the RNA problem. The detailed architectures are given in Figure~\ref{Architecture1forcontactmap1} and Figure~\ref{Architecture2forcontactmapSym}. 

We test our models on the protein sequences from the PDB25 list which is the dataset used in \cite{Wang_17AccurateDeNovo} and is  available at \url{http://dunbrack.fccc.edu/PISCES.php}. We test on a subset of length $25-100$ with $880$  training sequences, $220$ validation  and $73$ testing sequences.

\emph{Evaluation metrics:} We use the same metrics (PPV, Sensitivity, and Accuracy) defined for the  RNA problem in \S \ref{sec:rna}  to evaluate the performance of the models.

\emph{Implementation Details:} For the contact map prediction problem, we use the same implementation setup as in the RNA problem. We developed an architecture motivated by the RNA problem, that uses traditional convolution layers which we will refer to as the CNN model for contact map prediction. Figure~\ref{Architecture1forcontactmap1} presents our deep learning architecture for contact map prediction or a CNN model. Then we incorporate the symmetrical structure for the convolution layers, and we call it symmetrized CNN or SCNN for contact map prediction.

We observe that the contact map binary representation can be considered the label data for our protein sequence. In the labeled dataset, we use two-dimension representation using an array $A \in \mathbb{R}^{L \times L}$ where $L$ is the sequence length of primary protein structure. We let $C$ be the set of contact pairs, that is, $ C $ represents a pair of residues that have $C_{\alpha}$ atoms with Euclidean distance less than $8 \angstrom$, so $A^{(i,j)} = 1$ if $(i,j) \in C$ and $0$ otherwise. We can consider this two-dimensional representation as an image that is a natural fit for convolutional neural networks. We will use this representation as an output of the neural network. Instead of a binary matrix, entries of the network output $\hat Y$ will be bounded between 0 and 1, with $\hat Y^{(i,j)} \in [0,1]$ representing the predicted probability that $(i,j)$ has contact pairs in the contact map structure. We use these probabilities to appropriately convert the set of probabilistic contact predictions into a coherent contact map structure prediction.

We use the 2D array representation from the previous section to represent the output. Thus we presently have a problem with a 1D sequential input and a 2D sequential output. Like the RNA problem, we use a self-Cartesian product to obtain a 2D representation from a 1D representation of the features. The output of this Cartesian product is a 2D representation of the input where a square around the pixel $y^{(i, j)}$ contains local information about the residues near $x^{(i)}$ and $x^{(j)}$. To catch local features, we use convolutional layers; to identify global features, bidirectional LSTM is used.

As presented in Figure~\ref{Architecture1forcontactmap1}, we begin with the 1D input sequential data of size $L \times 25$. We first pass the sequence through a bidirectional LSTM to generate an output of size $L \times 15$. This output is concatenated with the 1D input data  $L \times 25$ to obtain a layer of $L \times 40$. We then perform the self-Cartesian product on $L \times 40$, making arrays of size $L \times L \times 80$. These are concatenated with the 2D input features of size $L \times L \times 3$ to obtain a layer of size $L \times L \times 83.$ These are passed through several 2D convolutional layers. We use several different kernel sizes at each layer to capture features of varying sizes without a large increase in the number of parameters. Also, it is essential to add the original features to the network again after many layers; in this case, we use the self-Cartesian of 1D features concatenated with 2D features. After concatenating original features, a few more 2D convolutional layers are used before applying the sigmoid layer to get probability values.

As presented in Figure~\ref{Architecture2forcontactmapSym}, we follow the same procedure with the CNN model for contact map prediction by attaching the symmetrical structure. We begin with the 1D input sequential data of size $L \times 25$. We first pass the sequence through a bidirectional LSTM to generate an output of size $L \times 15$. This output is concatenated with the 1D input data  $L \times 25$ to obtain a layer of $L \times 40$. We then perform the self-Cartesian product on $L \times 40$, making arrays of size $L \times L \times 80$. These are fed into the symmetry generating CNN layer with kernel size one to obtain the symmetric output, then concatenated with the 2D input features of size $L \times L \times 3$ to incorporate 2D input to the network. These are passed through symmetry preserving 2D convolutional layers. We use several different kernel sizes at each layer to capture features of varying sizes without a large increase in the number of parameters. Also, it is essential to add the original features to the network again after many layers. In this case, we use the self Cartesian of 1D features fed into the symmetry generating convolutional layer, then concatenate it with 2D features.  After concatenating symmetrized original features, a few more 2D convolutional layers are used before applying the sigmoid layer to get probability values. 

To train the parameters in the direct contact map prediction problem with the CNN and SCNN models, we used the Adam optimizer~\cite{kingma2014adam} with an initial learning rate of $10^{-3}$ for the CNN model while $2 \times 10^{-3}$ for the SCNN model. We use batch size $10$ for the network training. We also use an $L2$ regularization term of $10^{-5}$ and apply batch normalization as we indicated for the RNA problem before.  We use a weighted positive prediction term in the loss function using a small constant $5$.

We use the maximum sequence length to be $100$ for our experiments due to computing resource constraints; one can allow more lengthy sequences. Also, as the sizes of sequences in one particular batch may be incompatible, we find the maximum sequence length $L_0$ among the batch, and use value zero for the smaller sequence to match the data size of each batch sequence. Even though this batch generating method is used to handle the training process, in testing time, we can use the entire sequence individually, which does not affect the performance of the network.

\begin{table}[!ht]
        \centering
        \begin{tabular}{l|l|c|c}
            \toprule
            \textbf{Set} & \textbf{Metric} & \textbf{CNN} & \shortstack{\textbf{SCNN}
            } \\ 
            \midrule
            \multirow{3}{*}{Training}
            & PPV & 0.893 & \textbf{0.909}  \\%
            & Sen & 0.872 & \textbf{0.890}  \\%
            & Acc & 0.883 & \textbf{0.899}  \\ 
            \midrule
            \multirow{3}{*}{Validation}
            & PPV  & 0.841 & \textbf{0.895}  \\ 
            & Sen & 0.843 & \textbf{0.923}  \\ 
            & Acc & 0.842 & \textbf{0.909} \\ 
            \midrule
            \multirow{3}{*}{Test}
            & PPV & 0.952 & \textbf{0.981} \\ 
            & Sen & 0.934 & \textbf{0.967} \\
            & Acc & 0.943 & \textbf{0.974} \\
            \midrule
            \multicolumn{2}{r|}{\# of trainable param.} & 2,465,000 & 1,326,000 \\
            \bottomrule
        \end{tabular}
        \caption{Results on the Contact map prediction task using CNN and SCNN with hyperparameter matching model.}
        \label{ContactMapsymmetric_Nonsymmetric}
\end{table}

Table \ref{ContactMapsymmetric_Nonsymmetric} presents the results on the contact map prediction task using CNN and our SCNN models. Here the two models have the same  hyperparameters where we match the kernel size and feature sizes. Then, SCNN uses fewer trainable parameters. We note that our SCNN model outperforms the general CNN  architecture by $1.6\%$, $6.7\%$, and $3.1\%$ in training, validation, and testing accuracy respectively with over $45\%$ fewer parameters. We observe that in Table~\ref{ContactMapsymmetric_Nonsymmetric},  the outperformance is a bit stronger  in testing than in training and validation. This is partly due to batch training where sequences are augmented to have the same length in a mini-batch and all performance metrics are computed for the augmented sequences in training and validation, while individual protein sequences are used in test evaluation and we take the average over them. 

\section{Conclusion}
We have developed a new symmetrized convolutional neural network architecture, SCNN, to generate and maintain symmetric structures in CNN using symmetry generating CNNs and symmetry preserving CNNs. An update scheme to optimize trainable parameters used in the symmetry generating and preserving kernels was presented. We have demonstrated in three problems, where symmetric feature maps are desirable, that our SCNN architectures can improve  performance  with fewer trainable parameters, saving computational  costs at both training and inference.

\appendices
\section{Theory}\label{Appendix:A}

In this section, we present the proofs for the two theorems stated in section \ref{SSCNN}.
\subsubsection{Proof of Theorem \ref{SymmetryGeneratingThm}}
\setcounter{theorem}{0}
\begin{proof}
Note that $C$, $2n$, and $F$ are the kernel size, the number of input channels and the number of output channels. Let $\textbf{Z}=\textbf{W}*y$ be the output of the convolutional layer. We consider the $(s,t,f)^{\text{th}}$ entry of  $\textbf{Z}$: 

\begin{align*}
 &\textbf{Z}_{s,t,f}  = (\textbf{W}*y)_{s,t,f} \\
&=\sum_{k=1}^{n} \sum_{i,j=1}^C \textbf{W}_{i,j,k,f}x_{k}^{(s+i)}+\sum_{k=n+1}^{2n} \sum_{i,j=1}^C \textbf{W}_{i,j,k,f}x_{k-n}^{(t+j)}\\
&= \sum_{k=1}^{n} \sum_{i,j=1}^C \textbf{W}_{j,i,k,f}x_{k}^{(s+j)}+\sum_{k=1}^{n} \sum_{i,j=1}^C \textbf{W}_{j,i,k,f}x_{k}^{(t+i)}\\
&= \sum_{k=1}^{n} \sum_{i,j=1}^C \textbf{W}_{i,j,k,f}x_{k}^{(t+i)}+\sum_{k=1}^{n} \sum_{i,j=1}^C \textbf{W}_{i,j,k,f}x_{k}^{(s+j)}\\
&= \sum_{k=1}^{n} \sum_{i,j=1}^C \textbf{W}_{i,j,k,f}x_{k}^{(t+i)}+\sum_{k=n+1}^{2n} \sum_{i,j=1}^C \textbf{W}_{i,j,k,f}x_{k-n}^{(s+j)}\\
&= (\textbf{W}*y)_{t,s,f} = \textbf{Z}_{t,s,f}.
\end{align*}

\noindent As $\textbf{Z}_{s,t,f} = \textbf{Z}_{t,s,f}$ for any $f$, we have the symmetry of the output as desired.

\end{proof}

\subsubsection{Proof of Theorem \ref{SymmetryGeneratingupdateThm}}
\begin{proof}
For any  $i>j$, $k \leq n$, and  $f \leq F$, all the entries $\textbf{W}_{i,j,k,f}$, $\textbf{W}_{j,i,k,f}$, $\textbf{W}_{i,j,k+n,f}$, and $\textbf{W}_{j,i,k+n,f}$ are equal and are parameterized by $\textbf{S}_{i,j,k,f}$. Thus the gradient term for $i > j$ is

\begin{align}
\label{eq_symgenKernelUpdatesigj}
\displaystyle
\frac{\partial L}{\partial \textbf{S}_{i,j,k,f}}&=\displaystyle \sum_{s,t,u,v}\frac{\partial L}{\partial \textbf{W}_{s,t,u,v}} \frac{\partial \textbf{W}_{s,t,u,v}}{\partial \textbf{S}_{i,j,k,f}}\nonumber \\
&=\displaystyle \frac{\partial L}{\partial \textbf{W}_{i,j,k,f}}+ \frac{\partial L}{\partial \textbf{W}_{j,i,k,f}}\\
&\quad+\frac{\partial L}{\partial \textbf{W}_{i,j,k+n,f}}+\frac{\partial L}{\partial \textbf{W}_{j,i,k+n,f}}\nonumber,
\end{align}
where $\frac{\partial \textbf{W}_{s,t,u,v}}{\textbf{S}_{i,j,k,f}} = 0$ except indices $(i,j,k,f)$, $(j,i,k,f)$, $(i,j,k+n,f)$, and $(j,i,k+n,f)$.

For $i=j$,  $\textbf{W}_{i,i,k,f}$ and $ \textbf{W}_{i,i,k+n,f}$ are parameterized by $\textbf{S}_{i,i,k,f}$. This yield the gradient:

\begin{align}
\label{eq_symgenKernelUpdatesiequalj}
\displaystyle
\frac{\partial L}{\partial \textbf{S}_{i,i,k,f}}&=\displaystyle \sum_{s,t,u,v}\frac{\partial L}{\partial \textbf{W}_{s,t,u,v}} \frac{\partial \textbf{W}_{s,t,u,v}}{\partial \textbf{S}_{i,i,k,f}} \nonumber \\
&=\displaystyle \frac{\partial L}{\partial \textbf{W}_{i,i,k,f}}+ \frac{\partial L}{\partial \textbf{W}_{i,i,k+n,f}},
\end{align}
where $\frac{\partial \textbf{W}_{s,t,u,v}}{\textbf{S}_{i,i,k,f}} = 0$ for  $(s,t,u,v)\notin \{ (i,i,k,f), (i,i,k+n,f)\}$.
From Equation~(\ref{eq_symgenKernelUpdatesigj}) and Equation~(\ref{eq_symgenKernelUpdatesiequalj}) we have the desired gradient as in Equation (3) in the main text of the paper.

\end{proof}

\section*{Acknowledgment}
We thank Rebecca Calvert for reading the manuscript and providing us with many valuable comments/suggestions. We would also thank the University of Kentucky Center for Computational Sciences and Information Technology Services Research Computing for their support and use of the Lipscomb Compute Cluster and associated research computing resources.

\ifCLASSOPTIONcaptionsoff
  \newpage
\fi

\end{document}